\pdfoutput=1
\documentclass[11pt]{article}

\usepackage[]{coling}
\usepackage{amsmath} 
\usepackage{times}
\usepackage{latexsym}
\usepackage[utf8]{inputenc}
\usepackage{CJKutf8}
\usepackage{multirow} % Essential for creating multirow cells
\usepackage{nert}
\usepackage{array}    
\usepackage{graphicx}
\usepackage{adjustbox}
\usepackage{colortbl}
\usepackage{booktabs}

\usepackage{colortbl}
\definecolor{lightgray}{gray}{0.9}  % Adjust the shade of gray (0 to 1 scale)
\definecolor{beaublue}{rgb}{0.74, 0.83, 0.9}
\definecolor{airforceblue}{rgb}{0.36, 0.54, 0.66}
\usepackage{microtype}
\usepackage[T1]{fontenc}
\usepackage[utf8]{inputenc}

\usepackage{microtype}

\usepackage{inconsolata}
\usepackage{booktabs} 
\usepackage{graphicx}
\usepackage{expex}

\usepackage{graphicx}  % For \resizebox
\usepackage{booktabs}  % For \toprule, \midrule, etc.
\usepackage{rotating}  % For \rotatebox
\usepackage{colortbl}  % For coloring cells
\usepackage{lscape}    % If you need landscape orientation

\definecolor{blue1}{RGB}{255, 255, 255}   % White
\definecolor{blue2}{RGB}{198,219,239}  % Light blue
\definecolor{blue3}{RGB}{158,202,225}  % Medium light blue
\definecolor{blue4}{RGB}{107,174,214}  % Medium blue
\definecolor{blue5}{RGB}{49,130,189}   %   % Dark Blue  % Slightly Darker BlueYour specified color (Blue)

\newcommand{\colorcell}[1]{%
    \ifdim #1 pt > 80 pt \cellcolor{blue5}#1\else%
    \ifdim #1 pt > 60 pt \cellcolor{blue4}#1\else%
    \ifdim #1 pt > 40 pt \cellcolor{blue3}#1\else%
    \ifdim #1 pt > 20 pt \cellcolor{blue2}#1\else%
\cellcolor{blue1}#1\fi\fi\fi\fi%
}

\definecolor{gray1}{RGB}{242, 242, 242}  % White
\definecolor{gray2}{RGB}{230, 230, 230}   % Light gray
\definecolor{gray3}{RGB}{200, 200, 200}   % Medium light gray
\definecolor{gray4}{RGB}{150, 150, 150}   % Medium gray
\definecolor{gray5}{RGB}{100, 100, 100}   % Dark gray

\newcommand{\colorcellgray}[1]{%
    \ifdim #1 pt > 80 pt \cellcolor{gray5}#1\else%
    \ifdim #1 pt > 60 pt \cellcolor{gray4}#1\else%
    \ifdim #1 pt > 40 pt \cellcolor{gray3}#1\else%
    \ifdim #1 pt > 20 pt \cellcolor{gray2}#1\else%
    \cellcolor{gray1}#1\fi\fi\fi\fi%
}

\usepackage{array}

\title{Language Models at the Syntax-Semantics Interface:\\ A Case Study of the Long-Distance Binding of Chinese Reflexive \textit{ziji}}

\author{Xiulin Yang \\
  Georgetown University \\
  \texttt{xy236@georgetown.edu} \\}

\begin{document}
\maketitle
\begin{abstract}
This paper explores whether language models can effectively resolve the complex binding patterns of the Mandarin Chinese reflexive \textit{ziji}, which are constrained by both syntactic and semantic factors. We construct a dataset of 240 synthetic sentences using templates and examples from syntactic literature, along with 320 natural sentences from the BCC corpus. Evaluating 21 language models against this dataset and comparing their performance to judgments from native Mandarin speakers, we find that none of the models consistently replicates human-like judgments. The results indicate that existing language models tend to rely heavily on sequential cues, though not always favoring the closest strings, and often overlooking subtle semantic and syntactic constraints. They tend to be more sensitive to noun-related than verb-related semantics\footnote{Code and data are accessible via \url{https://github.com/xiulinyang/zh-reflexive}}. 

\end{abstract}
\section{Introduction}
Binding is a specific type of co-indexation that “lies at the very heart and soul of human language” \citep{abbott2010reference}. In a sentence, if a noun phrase  $NP_A$  binds another noun phrase  $NP_B$ , it indicates that both refer to the same entity \citep{carnie2021syntax}. In such cases, a pronoun or reflexive that refers back to an NP is called an \textit{anaphor}, and the NP it refers to is termed the \textit{antecedent}.

The impressive performance of Pretrained Language Models (PLMs) in various NLP tasks has raised an important question: \textbf{Do these models inherently acquire abstract linguistic knowledge solely from their training on sequences of strings?} To investigate this, many researchers treat language models as psycholinguistic objects. By designing minimal pairs and analyzing the probabilistic outputs from these models, they assess the models’ preferences in linguistic judgments.
 \begin{figure}[]
\centering
\includegraphics[width=\linewidth]{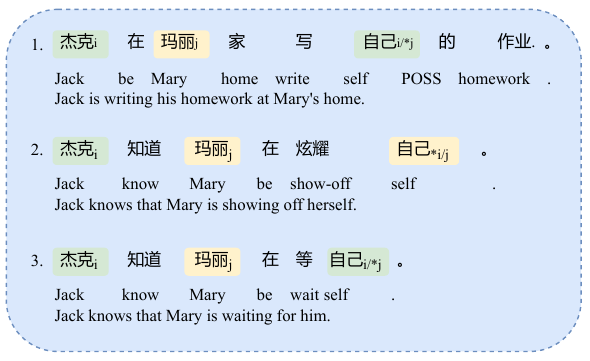}

    \caption{Examples of binding, the words highlighted in the same color are co-indexed.} 
    % p29/d2459
    \label{fig:example-binding}
\end{figure}
Numerous studies have explored linguistic patterns across different levels using minimal pairs, such as syntax \citep[e.g.,][]{wilcox-etal-2018-rnn,linzen2021syntactic,de-dios-flores-etal-2023-dependency}, semantics \citep[e.g.,][]{zhang-etal-2023-language}, and pragmatics \citep[e.g.,][]{davis-2022-incremental}. A few have examined the binding phenomenon in English, such as Reflexive Anaphor Licensing \citep{hu-etal-2020-closer,warstadt2020blimp, lee-schuster-2022-language,marvin-linzen-2018-targeted} and the constraints of Principle B in Binding Theory \citep{davis-2022-incremental}. There has also been work on binding in Chinese \citep{xiang-etal-2021-climp,song-etal-2022-sling}, but these studies typically focus on simple cases like gender/number agreement in local binding of complex reflexive \textit{ta-ziji} (\textit{himself/herself}).

However, binding in Chinese reflexives, particularly with the bare form \textit{ziji}, involves more than just gender agreement -- it is governed by intricate syntactic, semantic, and pragmatic constraints (see accounts from \citet{pan1998closeness,pan2003prominence, lam2021constraint}). Chomsky’s Binding Theory \citep{chomsky1993lectures}, especially Principle A, explains English reflexives but fails to generalize to long-distance reflexives like \textit{ziji} in Mandarin and others\footnote{e.g., \textit{zibun} in Japanese and \textit{kaki} in Teochew \citep{cole2006long}.} (see \ref{ldb}).  As shown in Figure~\ref{fig:example-binding}, different syntactic structures and verb types can lead to varied readings.

This study aims to investigate how language models handle the nuanced syntactic and semantic constraints in the complex binding patterns of the reflexive \textit{ziji} in Mandarin Chinese. Specifically, we seek to answer the following research questions:
\begin{itemize}
\item Can language models accurately process the intricate binding patterns of \textit{ziji} as humans do?
\item What factors contribute to the alignment or discrepancy between human judgments and the predictions made by these models?
\end{itemize}

We examine various language models, including monolingual, multilingual, masked language models, and autoregressive models, using both synthetic and natural datasets.

Our findings indicate that most models can predict some linguistic constraints but primarily rely on linear preferences rather than fully grasping syntax or semantics. Interestingly, not all models prefer binders closer to \textit{ziji}; for instance, \texttt{bert-base-chinese} favors long-distance binders. Furthermore, our experiments show that the models are better at noun-related semantics than verb-related ones.

Our key contributions are as follows: (1) We introduce a unique dataset comprising both synthetic and natural examples, along with human evaluation. To our knowledge, this is the first study to publicly provide human judgments and data in exploring the long-distance binding of \textit{ziji}; (2) we conduct one of the first comprehensive evaluations of multiple language models on the syntax-semantics interface in Chinese, revealing their limitations and investigating the underlying factors contributing to these shortcomings.

\section{Background \& Related Work}
\subsection{Chinese Reflexive \textit{ziji}}
Chinese reflexive \textit{ziji} has been extensively studied for decades due to its exceptional behaviors violating Principle A in the classic Binding Theory \citep{chomsky1993lectures}. According to Principle A, an anaphor must be bound within its governing category, typically the clause in which it appears. For example, in the sentence shown in Example (\ref{exp1}), the reflexive \textit{ziji} can only refer to the subject \textit{Mary}, following a pattern known as \textbf{local binding}.

\pex
\label{exp1}
\small
\begin{CJK}{UTF8}{gbsn}玛丽\textsubscript{j} 相信 自己\textsubscript{j} 。
\end{CJK} \\
\begingl %% Start glosses
\gla Mary\textsubscript{j} xiangxin ziji\textsubscript{j} //
\glb Mary trust self//
\glft \textit{Mary trusted herself}.//
\endgl
\xe

However, \textit{ziji} exhibits more complex behaviors beyond local binding. Its interpretation is governed by a range of syntactic and semantic constraints that have been extensively documented in the literature \citep[e.g.,][]{tang1989chinese,huang199113,pan2000blocking,charnavel2018inanimate,lam2021constraint,charnavel2018inanimate}. These complexities make \textit{ziji} a particularly intriguing case for studying reflexive binding patterns. In this research, we focus on several of these.

\paragraph{Long-distance Binding} \label{ldb} The antecedent can be bound remotely by the matrix subject in a complex clause \citep{liejiong1993long,huang199113,tang1989chinese}. For example, in (\ref{amb}), \textit{ziji} can refer to either the antecedent within the subordinate clause (\textit{Mary}) or beyond the clause (\textit{Jack}). 

\pex
\label{amb}
\small
\begin{CJK}{UTF8}{gbsn}杰克\textsubscript{i} 知道 玛丽\textsubscript{j} 相信 自己\textsubscript{i/j}。
\end{CJK} \\
\begingl %% Start glosses
\gla Jack\textsubscript{i} zhidao Mary\textsubscript{j} xiangxin ziji\textsubscript{i/j} //
\glb Jack knew Mary trust self//
\glft \textit{Jack knew that Mary trusted herself/him}.//
\endgl
\xe

\paragraph{Blocking Effect} In a complex clause, when the first/second person pronoun is inserted between the long-distance binder and reflexive, the long-distance binding is blocked or not allowed \citep{pan2000blocking}. Different from (\ref{amb}), in (\ref{bl}), \textit{ziji} instead can only refer to the first person \textit{I} but not \textit{Jack}.

\pex
\small
\label{bl}
\begin{CJK}{UTF8}{gbsn}杰克\textsubscript{i}知道我\textsubscript{j}相信自己\textsubscript{*i/j}。\end{CJK}\\
 \begingl %% Start glosses
\gla Jack\textsubscript{i} zhidao wo\textsubscript{j} xiangxin ziji\textsubscript{*i/j} //
\glb Jack knew I trust self//
\glft \textit{Jack knew that I trusted myself}.//
\endgl
\xe

\paragraph{Animacy Effect} The antecedent of \textit{ziji}
must be animate \citep{tang1989chinese}. In sentence (\ref{animacy}), although syntax allows both local binding and long-distance binding, the inanimacy of the subordinate subject makes the local binding impossible\footnote{Recent studies \citep[e.g.,][]{charnavel2018inanimate, lam2021constraint} have challenged the Animacy Effect assumption, but the counter-examples they provide are only limited to specific constructions which do not overlap with those used in our experiments.}.

\pex
\small
\label{animacy}
\begin{CJK}{UTF8}{gbsn}杰克\textsubscript{i}说这本书\textsubscript{j}欺骗了自己\textsubscript{i/*j}。\end{CJK}\\
 \begingl %% Start glosses
\gla Jack\textsubscript{i} shuo zhe ben shu\textsubscript{j} qipian le ziji\textsubscript{i/*j} //
\glb Jack say this CLS book deceive ASP self//
\glft \textit{Jack said that this book deceived him}.//
\endgl
\xe

\paragraph{Subject Orientation} Only the subject or part of the subject (e.g., possessor) can be a possible antecedent of \textit{ziji} \citep{tang1989chinese,lam2021constraint}. For example, in (\ref{subject_o}), only \begin{CJK}{UTF8}{gbsn}杰克\end{CJK} \textit{Jack} can be the antecedent of \textit{ziji} because it is the only subject in the clause. The other pronoun \begin{CJK}{UTF8}{gbsn}她\end{CJK} (\textit{her}) is the indirect object of the verb \begin{CJK}{UTF8}{gbsn}告诉\end{CJK} (\textit{tell}).
\pex
\label{subject_o}
\small
\begin{CJK}{UTF8}{gbsn}杰克\textsubscript{i}告诉她\textsubscript{j}自己\textsubscript{i/*j}的成绩。\end{CJK}\\
  \begingl %% Start glosses
\gla Jack\textsubscript{i} gaosu ta\textsubscript{j} ziji\textsubscript{i/*j} de chengji //
\glb Jack tell her self DE grade//
\glft \textit{Jack told her his grade}.//
\endgl
\xe
\paragraph{Verb Orientation} Studies have also found that in complex sentences, the meaning of subordinate predicates might disambiguate the possible readings of \textit{ziji} \citep{qiu, schumacher2011perspective}. For instance, the following two examples share the same syntactic structures but have different readings because of the semantics of the subordinate predicate. For (\ref{ov}), the flatterer typically flatters someone else rather than themselves. By contrast, in (\ref{iv}), one can only reflect on their own mind, not others.

\pex
\a 
\label{ov}
\small
\begin{CJK}{UTF8}{gbsn}杰克\textsubscript{i}说玛丽\textsubscript{j}巴结了自己\textsubscript{i/*j}。\end{CJK}\\
  \begingl 
\gla Jack\textsubscript{i} shuo Mary\textsubscript{j} bajie le ziji\textsubscript{i/*j} //
\glb Jack say Mary flatter ASP self//
\glft \textit{Jack said that Mary flattered him.}//
\endgl
\a 
\label{iv}
\begin{CJK}{UTF8}{gbsn}杰克\textsubscript{i}说玛丽\textsubscript{j}反省了自己\textsubscript{*i/j}。\end{CJK}\\
    \begingl %% Start glosses
\gla Jack\textsubscript{i} shuo Mary\textsubscript{j} fanxing le ziji\textsubscript{*i/j} //
\glb Jack say Mary reflect.on ASP self//
\glft \textit{Jack said that Mary reflected on herself.}//
\endgl
\xe

\subsection{Probing Linguistic Knowledge in Language Models}
To examine linguistic knowledge in language models, many studies have explored syntactic and semantic structures such as subject-verb agreement \citep{marvin-linzen-2018-targeted}, Negative Polarity Items \citep{jumelet-etal-2021-language}, and long-distance dependencies \citep{marvin-linzen-2018-targeted} across languages \citep{xiang-etal-2021-climp,de-dios-flores-etal-2023-dependency}. Findings show that while models produce syntactically correct output, their performance may be biased by dependency distance or token frequency \citep{newman-etal-2021-refining,wei-etal-2021-frequency}.

Probing methods in NLP serve as crucial tools for deciphering the intricate workings of language models. These methods range from probing tasks, which assess the model's grasp on linguistic properties through external classifiers 
\citep[e.g.,][]{wu-dredze-2019-beto, levy-goldberg-2014-linguistic, tenney-etal-2019-bert,kulmizev-etal-2020-neural}, to attention analysis \citep{voita-etal-2019-analyzing}, aimed at understanding focus patterns within the network. Further, many studies employ concepts from Information Theory such as perplexity and surprisal to investigate the linguistic behaviors of language models by taking them as psycholinguistic objects \citep{futrell-etal-2019-neural, wilcox2023testing, oh2022comparison}. Recently, as more LLMs shift towards closed-source, researchers have turned to prompting techniques to explore their knowledge \citep{katzir2023large,Ambridge2024LargeLM,Lamprinidis2023LLMCJ,Dentella2023TestingAO}. In our study, we employ the perplexity-based method for open-source models and prompting for closed-source models.

Most research on language models has focused on languages whose case and agreement systems facilitate controlled experiments \citep{de-dios-flores-etal-2023-dependency}. In contrast, less focus has been given to the syntax-semantics interface in morphologically poor languages like Mandarin Chinese. \citet{de-dios-flores-etal-2023-dependency} found that language models often mispredict anaphora resolution in Spanish and Galician when antecedents and anaphors are distantly placed. Similar trends have been observed in English studies \citep{lee-schuster-2022-language}. While some research has examined Chinese linguistic knowledge in models like BERT, it has mainly addressed syntax \citep{zheng2023does,kulmizev-etal-2020-neural,xiang-etal-2021-climp}, leaving the syntax-semantics interface largely unexplored. Our study aims to fill this gap by investigating the binding of \textit{ziji} in Mandarin Chinese.
 
\begin{table*}[!ht]
\small
    \centering
    \begin{tabular}{c|c|c|c|c} 
         \toprule
         \textbf{Binding Pattern} & \textbf{Constraint} & \textbf{Categories} & \textbf{Example} & \textbf{Gold Binding}\\ \toprule
         \multirow{2}{*}{\parbox{2cm}{Ambiguous Long distance binding (AMB LD)}} & \multirow{2}{*}{Syntax\&Semantics} & Fem pronoun first &  \parbox{5cm}{\vspace{0.3cm}\begin{CJK}{UTF8}{gbsn}她\textsubscript{f}知道他\textsubscript{m}相信自己\textsubscript{f/m}。\end{CJK} \\ \textit{She\textsubscript{f} knows that he\textsubscript{m} trusts himself\textsubscript{m}/her\textsubscript{f}.}\\} & Ambiguous \\ \cline{3-5}
         & & Masc pronoun first &   \parbox{5cm}{\vspace{0.3cm}\begin{CJK}{UTF8}{gbsn}他\textsubscript{m}知道她\textsubscript{f}相信自己\textsubscript{m/f}。\end{CJK} \\ \textit{He\textsubscript{m} knows that she\textsubscript{f} trusts herself\textsubscript{f}/him\textsubscript{m}.}\\} &  Ambiguous\\ \hline
         
         \multirow{2}{*}{\parbox{2cm}{Verb Orientation (VO)}} & \multirow{2}{*}{Semantics} & Reflexive &  \parbox{5cm}{\vspace{0.3cm}\begin{CJK}{UTF8}{gbsn}她\textsubscript{f}知道他\textsubscript{m}在检讨自己\textsubscript{m}。\end{CJK} \\ \textit{She\textsubscript{f} knows that he\textsubscript{m} is reflecting on himself\textsubscript{m}.}\\} & Local \\ \cline{3-5}
         & & No-reflexive &   \parbox{5cm}{\vspace{0.3cm}\begin{CJK}{UTF8}{gbsn}他\textsubscript{m}知道她\textsubscript{f}在躲避自己\textsubscript{m}。\end{CJK} \\ \textit{He\textsubscript{m} knows that she\textsubscript{f} is escaping him\textsubscript{m}.}\\} & Remote\\ \hline
         \multirow{2}{*}{\parbox{2cm}{Subject Orientation (SO)}} & \multirow{2}{*}{Syntax} & Fem pronoun first &  \parbox{5cm}{\vspace{0.3cm}\begin{CJK}{UTF8}{gbsn}她\textsubscript{f}给他\textsubscript{m}关于自己\textsubscript{f}的书。\end{CJK} \\ \textit{She\textsubscript{f} gave him\textsubscript{m} her\textsubscript{f} own book.}\\} & Remote \\ \cline{3-5}
         & & Masc pronoun first &   \parbox{5cm}{\vspace{0.3cm}\begin{CJK}{UTF8}{gbsn}他\textsubscript{m}给她\textsubscript{f}关于自己\textsubscript{f}的书。\end{CJK} \\ \textit{He\textsubscript{m} gave her\textsubscript{f} his\textsubscript{m} own book.}\\} & Remote\\ \hline

         \parbox{2cm}{Blocking Effect (BE)} & Syntax\&Semantics & NA &  \parbox{5cm}{\vspace{0.3cm}\begin{CJK}{UTF8}{gbsn}她\textsubscript{f}知道我\textsubscript{m}相信自己\textsubscript{m}。\end{CJK} \\ \textit{She\textsubscript{f} knows that I\textsubscript{w} trust myself\textsubscript{w}.}\\} & Local \\ \hline

        \parbox{2cm}{Animacy Effect (AE)} & Semantics & NA &  \parbox{5cm}{\vspace{0.3cm}\begin{CJK}{UTF8}{gbsn}她\textsubscript{f}知道这封信\textsubscript{t}暴露了自己\textsubscript{f}。\end{CJK} \\ \textit{She\textsubscript{f} knows that the letter\textsubscript{t} exposes her\textsubscript{f}.}\\} & Remote \\ \hline
         
    \end{tabular}
    \caption{Binding patterns and their corresponding examples. Among the subscript following NPs, \textit{m} refers to third person masculine pronoun, \textit{f} refers to third person feminine pronoun, \textit{w} refers to the first-person pronoun, \textit{t} refers to the third person inanimate pronoun. \textit{Remote} means the antecedent is linearly farther away than the incorrect binder or distractor, not strictly \textit{long-distance} binding explained in \ref{ldb}.}
    \label{tab:examples}
\end{table*}

\section{Data} 
To address potential discrepancies between synthetic and natural data used for training language models, we develop two distinct datasets: one generated automatically via a script or hand-crafted by linguists, and the other collected from the BCC corpus\footnote{\url{https://bcc.blcu.edu.cn/}} \citep{XunEndong2016BigData}.

\subsection{Synthetic Data}
To create synthetic data, we choose the syntactic structure consistent with most psycholinguistic studies on long-distance binding of \textit{ziji} \citep[e.g.,][]{schumacher2011perspective,li2009overcoming}, i.e., \(NP_1 + V_1 + NP_2 + V_2 + \textit{ziji}\). This structure is used to test all constraints except for Subject Orientation. Given the focus of our experiments, we specifically varied \(NP_1\), \(NP_2\), \(V_1\), and \(V_2\). Due to the absence of morphological inflections in Chinese to mark agreement between antecedents and anaphors, we limited our selection of NP$_1$s to four single-character pronouns with different semantic features: \begin{CJK}{UTF8}{gbsn}他\end{CJK} \textit{(he/him)}, \begin{CJK}{UTF8}{gbsn}她\end{CJK} \textit{(she/her)}, \begin{CJK}{UTF8}{gbsn}我\end{CJK} \textit{(I/me)}, and \begin{CJK}{UTF8}{gbsn}它\end{CJK} \textit{(it)}.

\(V_1\) is always a statement/attitude verb. Regarding the choice of \(V_2\), we leveraged the comprehensive analysis by \citet{qiu}, who examined how the semantic properties of verbs influence the binding of \textit{ziji}. After reviewing the Chinese Verb Usage Dictionary, \citet{qiu} categorized verbs into three main types: non-reflexive verbs, which inhibit local binding of \textit{ziji} (e.g., sentence~(\ref{ov})); reflexive verbs, which prevent long-distance binding of \textit{ziji} (e.g., sentence~(\ref{iv})); and bidirectional verbs, which allow ambiguous interpretations of \textit{ziji} (e.g., sentence~(\ref{amb})). We select a random subset of verbs from the former two categories in our study to build sentence pairs for Verb Orientation tests.

Regarding the Subject Orientation constraint, we utilize the following two syntactic structures: \(NP_1 + V_1 + NP_2 + \textit{ziji} + de + NP_3\), where \(V_1\) is a ditransitive verb, \(NP_2\) is the indirect object, and \textit{ziji} serves as the possessor of the entire direct object phrase (i.e., \textit{ziji de $NP_3$ (one's own $NP_3$)}); and \(NP_1 + PP + V_1 + \textit{ziji} + de + NP_3\), where a distractor noun is inserted into the PP. In both cases, \textit{ziji} can only be bound by \(NP_1\).

As for the Blocking Effect, psycholinguistic studies have noted that this constraint is not absolute \citep{Lyu2021UnpackingTB}. To minimize potential biases arising from our templates or verb selection, we supplemented our dataset with 40 sentences from existing literature \citep{li2023makes, shuai2013interpretation, pan2000blocking, schumacher2011perspective, huang2002distributivity, chen2009logophoricity, liu2010pragmatic, yang2015whether, cole1994head}. 

Additionally, by replacing the first-person pronoun in sentences from the Blocking Effect category, we design 40 sentence pairs with the third-person pronoun to allow ambiguous binding and test language models' structural bias as a baseline. We aim to assess which binding -- local or long-distance -- language models/humans prefer when both are acceptable.

Overall, our experimental dataset comprises 240 sentences, with each category containing 40 examples. The linguistic patterns, example sentences, and correct binding are detailed in Table~\ref{tab:examples}.

\subsection{In-Context Minimal Pairs}
We design the synthetic dataset to test the reading of \textit{ziji}. However, we cannot get who \textit{ziji} refers to simply from the sequence because \textit{ziji} itself does not have any morphological cue to indicate its binder. To address this, we develop a method we call \textit{in-context minimal pairs}. We embed the target sentence in a structure like: \textit{If [TARGET SENTENCE], then [INTERPRETATION OF TARGET SENTENCE]}. In the second clause, the semantic feature of the binder is made explicit by using pronouns or complex reflexive (e.g., ta-ziji \textit{himself}). This approach allows us to test language models' preferred reading in a more natural context.  For instance, sentence (\ref{amb}) can be reformulated into the following minimal pair. (\ref{minimalpaira}) suggests a local binding of \textit{ziji}, while (\ref{minimalpairb}) suggests a long-distance binding. The minimal pair examples for different binding patterns are detailed in Appendix~\ref{template}.

\pex
\label{minimalpair}
\a
\label{minimalpaira}
\small
\begin{CJK}{UTF8}{gbsn}如果 杰克\textsubscript{i} 知道 玛丽\textsubscript{j} 相信 自己\textsubscript{i/j}， 那么 玛丽 相信 \textcolor{airforceblue}{\textbf{\textit{她自己}}}。
\end{CJK} \\
\begingl %% Start glosses
\gla if Jack\textsubscript{i} zhidao Mary\textsubscript{j} xiangxin ziji\textsubscript{i/j}, namo Mary xiangxin taziji //
\glb  if Jack knew Mary trust self,  then Mary trust herself.//
\glft \textit{If Jack knew that Mary trusted herself/him, then Mary trusted \textcolor{airforceblue}{herself}}.//
\endgl
\a
\label{minimalpairb}
\begin{CJK}{UTF8}{gbsn}如果 杰克\textsubscript{i} 知道 玛丽\textsubscript{j} 相信 自己\textsubscript{i/j}， 那么 玛丽 相信 \textcolor{airforceblue}{\textbf{\textit{他}}}。
\end{CJK} \\
\begingl %% Start glosses
\gla If Jack\textsubscript{i} zhidao Mary\textsubscript{j} xiangxin ziji\textsubscript{i/j}, namo Mary xiangxin ta. //
\glb if Jack knew Mary trust self, then Mary trust him//
\glft \textit{Jack knew that Mary trusted herself/him, then Mary trusted \textcolor{airforceblue}{him}}.//
\endgl
\xe

\subsection{Natural Data}
Since previous research has indicated that language models significantly underperform humans on reflexive binding tasks \citep{song-etal-2022-sling}, we aim to extend this evaluation to natural data to determine if similar conclusions hold. We manually select 240 natural sentences from the BCC corpus, ensuring they align with the structure of the synthetic data. Additionally, we collect 80 sentences specifically involving local binding in contrast with \citet{song-etal-2022-sling}'s data.
\begin{table}[]
\centering
    \small
    \resizebox{\columnwidth}{!}{ 
    \begin{tabular}{lccc}
    \toprule
      Model   & \# Params & Training Data Size \\
    \midrule
      bert-base-chinese \citep{devlin-etal-2019-bert}          & 110M   & 300G  \\
      chinese-lert-base \citep{cui2022lert}     & 102M   & 20GB  \\
      chinese-lert-large \citep{cui2022lert}   & 325M   & 20GB  \\
      chinese-pert-base \citep{cui2022pert}    & 102M   & 20GB  \\
      chinese-pert-large \citep{cui2022pert}   & 325M   & 20GB  \\
      mengzi-bert-base \citep{zhang2021mengzi}     & 103M   & 300G  \\
      mengzi-bert-base-fin \citep{zhang2021mengzi}  & 103M   & 320G  \\
      ernie-1.0-base-zh  \citep{sun2019ernie}   & 110M   & 173M sent. \\
      \rowcolor{beaublue}mBERT \citep{devlin-etal-2019-bert}               & 110M   & -     \\
      \rowcolor{beaublue}XLM-R-base  \citep{Conneau2019UnsupervisedCR}          & 125M   & 2.5TB \\
      \rowcolor{beaublue}XLM-R-large \citep{Conneau2019UnsupervisedCR}         & 355M   & 2.5TB \\
      \midrule
      \midrule
      \rowcolor{beaublue}mt5-small\citep{Xue2020mT5AM}            & 300M   & 0.5TB \\  % Added a hypothetical value for illustration
      \rowcolor{beaublue} mt5-large  \citep{Xue2020mT5AM}            & 1.2B   & 1TB    \\  % Added a hypothetical value for illustration
      \midrule
      \midrule
      GPT2 \citep{zhao2019uer}                & 117M   & 14GB  \\
      GPT2-medium  \citep{zhao2019uer}          & 345M   & 14GB  \\
      GPT2-large  \citep{zhao2019uer}           & 762M   & 14GB  \\
      GPT2-xlarge \citep{zhao2023tencentpretrain}         & 1.5B   & 14GB  \\
      \rowcolor{beaublue}GLM-4-9b-chat \citep{glm2024chatglm}        & 9B     & -     \\
      CPM-Generate \citep{zhang2021cpm}        & 2.6B   & 100GB \\
      \rowcolor{beaublue}GPT-3.5 \citep{openai2023gpt35} & NA & NA\\
      \rowcolor{beaublue}GPT-4o \citep{openai2023gpt4} & NA & NA\\
         \bottomrule 
    \end{tabular}}
    
    \caption{Overview of the models used in the experiments, categorized by architecture: encoder-only models, encoder-decoder models, and decoder-only models. The table also includes the corresponding number of parameters and training data sizes for each model. Multilingual models are highlighted in blue for clarity.}
    \label{model-parm}
 
\end{table}

For local binding constructions, we select sentences where \begin{CJK}{UTF8}{gbsn}她自己\end{CJK} (\textit{herself}) or \begin{CJK}{UTF8}{gbsn}他自己\end{CJK} (\textit{himself}) appears as the direct object. For other binding constructions except for Subject Orientation, we focus on sentences following the \(NP_1 + V_1 + NP_2 + V_2 + \textit{ziji}\) pattern, allowing for additional contextual elements or modifiers. For Subject Orientation, we select sentences that contain a distractor NP with a different gender feature between the antecedent and \textit{ziji}. To minimize potential confounds brought by gender bias in our experiments, we make minimal alterations to ensure gender balance in the dataset. 

Embedding natural sentences into an \textit{if ... then} template (or using other similar connectives) often makes them sound unnatural, as these sentences are longer than typical conditional clauses. This makes it difficult, if not impossible, to create natural-sounding in-context minimal pairs. We assume that using complex reflexives like \begin{CJK}{UTF8}{gbsn}她/他/我自己\end{CJK} (\textit{her/him/myself}) serves as a useful proxy for testing language models' preferred readings of \textit{ziji}, as the pronoun makes the reference explicit. Thus, we use minimal pairs by replacing \textit{ziji} with \textit{ta-ziji} to clarify its meaning where contextually appropriate.\footnote{See Appendix \ref{natural} for examples.} The gender and animacy features of the incorrect candidate are determined by the non-antecedent noun.

\begin{table*}[th]
\centering
\resizebox{\textwidth}{!}{%
\begin{tabular}{lcccccccccccccccccccccc}
& \rotatebox{75}{Human} & \rotatebox{75}{bert-base-chinese} & \rotatebox{75}{chinese-lert-base} & \rotatebox{75}{chinese-lert-large} & \rotatebox{75}{chinese-pert-base} & \rotatebox{75}{chinese-pert-large} & \rotatebox{75}{mengzi-bert-base} & \rotatebox{75}{mengzi-bert-base-fin} & \rotatebox{75}{ernie-1.0-base-zh} & \rotatebox{75}{multilingual-bert} & \rotatebox{75}{xlmr-base} & \rotatebox{75}{xlmr-large} & \rotatebox{75}{mt5-small} & \rotatebox{75}{mt5-large} & \rotatebox{75}{gpt2-distil} & \rotatebox{75}{gpt2-medium} & \rotatebox{75}{gpt2-chinese} & \rotatebox{75}{gpt2-xlarge} & \rotatebox{75}{glm-4-9b-chat} & \rotatebox{75}{CPM-Generate} & \cellcolor{lightgray}\rotatebox{75}{GPT-3.5} & \cellcolor{lightgray}\rotatebox{75}{GPT-4o} \\

\midrule
Blocking & \colorcell{92.5} & \colorcell{22.5} & \colorcell{10.0} & \colorcell{17.5} & \colorcell{15.0} & \colorcell{5.0} & \colorcell{42.5} & \colorcell{45.0} & \colorcell{15.0} & \colorcell{0.0} & \colorcell{7.5} & \colorcell{17.5} & \colorcell{17.5} & \colorcell{30.0} & \colorcell{100.0} & \colorcell{67.5} & \colorcell{100.0} & \colorcell{90.0} & \colorcell{75.0} & \colorcell{100.0} & \colorcell{62.5}& \colorcell{27.5} \\
Animacy & \colorcell{100.0} & \colorcell{100.0} & \colorcell{100.0} & \colorcell{100.0} & \colorcell{100.0} & \colorcell{75.0} & \colorcell{97.5} & \colorcell{97.5} & \colorcell{100.0} & \colorcell{100.0} & \colorcell{100.0} & \colorcell{100.0} & \colorcell{25.0} & \colorcell{27.5} & \colorcell{100.0} & \colorcell{100.0} & \colorcell{92.5} & \colorcell{100.0} & \colorcell{22.5} & \colorcell{100.0} & \colorcell{100.0}& \colorcell{100.0}\\
Verb$_{refl}$ & \colorcell{100.0} & \colorcell{5.0} & \colorcell{7.5} & \colorcell{2.5} & \colorcell{0.0} & \colorcell{37.5} & \colorcell{70.0} & \colorcell{45.0} & \colorcell{2.5} & \colorcell{52.5} & \colorcell{2.5} & \colorcell{20.0} & \colorcell{10.0} & \colorcell{20.0} & \colorcell{97.5} & \colorcell{95.0} & \colorcell{100.0} & \colorcell{100.0} & \colorcell{52.5} & \colorcell{5.0}& \colorcell{55.0}& \colorcell{65.0} \\
Verb$_{nonrefl}$ & \colorcell{97.5} & \colorcell{100.0} & \colorcell{100.0} & \colorcell{100.0} & \colorcell{100.0} & \colorcell{75.0} & \colorcell{62.5} & \colorcell{65.0} & \colorcell{100.0} & \colorcell{52.5} & \colorcell{100.0} & \colorcell{95.0} & \colorcell{92.5} & \colorcell{100.0} & \colorcell{7.5} & \colorcell{0.0} & \colorcell{0.0} & \colorcell{0.0} & \colorcell{100.0} & \colorcell{72.5}&\colorcell{100.0} & \colorcell{97.5}\\
SO & \colorcell{87.5} & \colorcell{37.5} & \colorcell{80.0} & \colorcell{87.5} & \colorcell{65.0} & \colorcell{52.5} & \colorcell{85.0} & \colorcell{50.0} & \colorcell{65.0} & \colorcell{7.5} & \colorcell{70.0} & \colorcell{45.0} & \colorcell{0.0} & \colorcell{22.5} & \colorcell{50.0} & \colorcell{47.5} & \colorcell{52.5} & \colorcell{60.0} & \colorcell{35.0} & \colorcell{47.5} &\colorcell{50.0} & \colorcell{50.0}\\
Average & \colorcell{95.5} & \colorcell{53.0} & \colorcell{59.5} & \colorcell{61.5} & \colorcell{56.0} & \colorcell{49.0} & \colorcell{71.5} & \colorcell{60.5} & \colorcell{56.5} & \colorcell{42.5} & \colorcell{56.0} & \colorcell{55.5} & \colorcell{29.0} & \colorcell{40.0} & \colorcell{71.0} & \colorcell{62.0} & \colorcell{69.0} & \colorcell{70.0} & \colorcell{57.0} & \colorcell{65.0} & \colorcell{65.0} & \colorcell{68.5} \\

\midrule
\midrule
Ambiguous & \colorcellgray{17.5} & \colorcellgray{15.0} & \colorcellgray{15.0} & \colorcellgray{12.5} & \colorcellgray{5.0} & \colorcellgray{22.5} & \colorcellgray{12.5} & \colorcellgray{5.0} & \colorcellgray{17.5} & \colorcellgray{20.0} & \colorcellgray{10.0} & \colorcellgray{20.0} & \colorcellgray{12.5} & \colorcellgray{87.5} & \colorcellgray{82.5} & \colorcellgray{95.0} & \colorcellgray{87.5} & \colorcellgray{60.0} & \colorcellgray{35.0} & \colorcellgray{50.0} & \colorcellgray{22.5} & \colorcellgray{35.0} \\

\bottomrule
\end{tabular}%
}
\caption{\small \textbf{Accuracy Scores} of Predictions on Synthetic Data and \textbf{Local Binding Percentage} on the Ambiguous setting (last row). Cells are shaded to reflect performance levels, with darker shades indicating higher accuracy. \textbf{Blocking} refers to the blocking effect setting; \textbf{Animacy} refers to the animacy effect experiment. \textbf{Verb$_{refl}$} refers to the reflexive subcategory within the Verb Orientation category, while \textbf{Verb$_{nonrefl}$} denotes the non-reflexive category. \textbf{SO} indicates Subject Orientation. The two gray-shaded GPT models are highlighted because they are evaluated using prompting rather than perplexity.}
\label{result_syn}
\end{table*}

\begin{table*}[th]
\centering
\resizebox{\textwidth}{!}{%
\begin{tabular}{lcccccccccccccccccccccc}
& \rotatebox{75}{Human} & \rotatebox{75}{bert-base-chinese} & \rotatebox{75}{chinese-lert-base} & \rotatebox{75}{chinese-lert-large} & \rotatebox{75}{chinese-pert-base} & \rotatebox{75}{chinese-pert-large} & \rotatebox{75}{mengzi-bert-base} & \rotatebox{75}{mengzi-bert-base-fin} & \rotatebox{75}{ernie-1.0-base-zh} & \rotatebox{75}{multilingual-bert} & \rotatebox{75}{xlmr-base} & \rotatebox{75}{xlmr-large} & \rotatebox{75}{mt5-small} & \rotatebox{75}{mt5-large} & \rotatebox{75}{gpt2-distil} & \rotatebox{75}{gpt2-medium} & \rotatebox{75}{gpt2-chinese} & \rotatebox{75}{gpt2-xlarge} & \rotatebox{75}{glm-4-9b-chat} & \rotatebox{75}{CPM-Generate} & \cellcolor{lightgray}\rotatebox{75}{GPT-3.5} & \cellcolor{lightgray}\rotatebox{75}{GPT-4o} \\
\midrule
Blocking & \colorcell{100} & \colorcell{50.0} & \colorcell{72.5} & \colorcell{72.5} & \colorcell{55.0} & \colorcell{52.5} & \colorcell{72.5} & \colorcell{72.5} & \colorcell{80.0} & \colorcell{55.0} & \colorcell{45.0} & \colorcell{65.0} & \colorcell{50.0} & \colorcell{90.0} & \colorcell{62.5} & \colorcell{47.5} & \colorcell{57.5} & \colorcell{55.0} & \colorcell{90.0} & \colorcell{72.5} & \colorcell{57.5} & \colorcell{100.0} \\
Animacy & \colorcell{97.5} & \colorcell{97.5} & \colorcell{97.5} & \colorcell{97.5} & \colorcell{100.0} & \colorcell{80.0} & \colorcell{97.5} & \colorcell{100.0} & \colorcell{97.5} & \colorcell{90.0} & \colorcell{82.5} & \colorcell{87.5} & \colorcell{15.0} & \colorcell{97.5} & \colorcell{97.5} & \colorcell{92.5} & \colorcell{97.5} & \colorcell{97.5} & \colorcell{75.0} & \colorcell{100.0} & \colorcell{80.0} & \colorcell{97.5}\\
Verb$_{refl}$ & \colorcell{100} & \colorcell{72.5} & \colorcell{62.5} & \colorcell{80.0} & \colorcell{72.5} & \colorcell{50.0} & \colorcell{65.0} & \colorcell{77.5} & \colorcell{65.0} & \colorcell{70.0} & \colorcell{70.0} & \colorcell{87.5} & \colorcell{57.5} & \colorcell{82.5} & \colorcell{45.0} & \colorcell{42.5} & \colorcell{32.5} & \colorcell{57.5} & \colorcell{92.5} & \colorcell{22.5} & \colorcell{60.0} & \colorcell{100.0}\\
Verb$_{nonrefl}$ & \colorcell{100} & \colorcell{75.0} & \colorcell{72.5} & \colorcell{95.0} & \colorcell{65.0} & \colorcell{85.0} & \colorcell{65.0} & \colorcell{72.5} & \colorcell{72.5} & \colorcell{30.0} & \colorcell{82.5} & \colorcell{77.5} & \colorcell{47.5} & \colorcell{87.5} & \colorcell{55.0} & \colorcell{80.0} & \colorcell{65.0} & \colorcell{82.5} & \colorcell{95.0} & \colorcell{75.0} & \colorcell{90.0} & \colorcell{100.0}\\
SO & \colorcell{97.5} & \colorcell{72.5} & \colorcell{80.0} & \colorcell{90.0} & \colorcell{57.5} & \colorcell{57.5} & \colorcell{87.5} & \colorcell{80.0} & \colorcell{72.5} & \colorcell{60.0} & \colorcell{70.0} & \colorcell{80.0} & \colorcell{27.5} & \colorcell{62.5} & \colorcell{55.0} & \colorcell{72.5} & \colorcell{65.0} & \colorcell{82.5} & \colorcell{85.0} & \colorcell{47.5} & \colorcell{60.0} & \colorcell{97.5}\\
Average & \colorcell{99.0} & \colorcell{73.5} & \colorcell{77.0} & \colorcell{87.0} & \colorcell{70.0} & \colorcell{65.0} & \colorcell{77.5} & \colorcell{80.5} & \colorcell{77.5} & \colorcell{61.0} & \colorcell{70.0} & \colorcell{79.5} & \colorcell{39.5} & \colorcell{84.0} & \colorcell{63.0} & \colorcell{67.0} & \colorcell{63.5} & \colorcell{75.0} & \colorcell{87.5} & \colorcell{63.5} & \colorcell{69.5} & \colorcell{99}\\
\midrule
\midrule
Local$_{m}$ & \colorcell{100} & \colorcell{85.0} & \colorcell{87.5} & \colorcell{90.0} & \colorcell{87.5} & \colorcell{10.0} & \colorcell{90.0} & \colorcell{87.5} & \colorcell{87.5} & \colorcell{82.5} & \colorcell{62.5} & \colorcell{90.0} & \colorcell{17.5} & \colorcell{77.5} & \colorcell{85.0} & \colorcell{87.5} & \colorcell{87.5} & \colorcell{97.5} & \colorcell{95.0} & \colorcell{72.5} & \colorcell{62.5} & \colorcell{92.5}\\
Local$_{f}$ & \colorcell{100} & \colorcell{85.0} & \colorcell{85.0} & \colorcell{92.5} & \colorcell{92.5} & \colorcell{87.5} & \colorcell{100.0} & \colorcell{100.0} & \colorcell{95.0} & \colorcell{82.5} & \colorcell{97.5} & \colorcell{92.5} & \colorcell{17.5} & \colorcell{80.0} & \colorcell{97.5} & \colorcell{95.0} & \colorcell{97.5} & \colorcell{95.0} & \colorcell{82.5} & \colorcell{80.0}& \colorcell{57.5} & \colorcell{97.5}\\

\bottomrule
\end{tabular}%
}
\caption{\small Accuracy Scores of Predictions on Natural Data. Cells are shaded to indicate performance levels, with darker shades representing higher accuracy scores. The last two rows show the results of \textbf{local binding} on two gender settings in natural data.}
\label{result_natural}
\end{table*}

\section{Evaluation}
\subsection{Models}
Following \citet{song-etal-2022-sling}, we evaluated most models used in their study and included additional language models trained on monolingual or multilingual corpora, featuring different architectures and sizes. In total, we examined 21 language models, including the latest \texttt{GPT-4o}. The number of parameters and the size of the training data can be found in Table~\ref{model-parm}.\footnote{Note that discrepancies from \cite{song-etal-2022-sling} may arise from references to different sources about the model information.}

\subsection{(Pseudo-) Perplexity}
In line with \citet{song-etal-2022-sling}, we evaluate the performance of autoregressive language models using perplexity (PPL) and masked language models using pseudo-perplexity (PPPL) \citep{salazar-etal-2020-masked}. The equations for PPL are defined as follows.

\begin{equation}
\begin{aligned}
&L = \frac{1}{M} \sum_{i=1}^{m} \log p(w_i | w_{1} \ldots w_{i-1}) \\
&\text{PPL} = \exp(-L)
\end{aligned}
\end{equation}

While PPL measures the probability of tokens based solely on preceding context, PPPL calculates the probability of a token using the entire bidirectional context, informed by the pretrained tasks of MLMs. 

\begin{equation}
\begin{aligned}
&w_{\backslash i} = w_{1} \ldots w_{i-1}, w_{i+1} \ldots w_{m} \\
&\text{pseudo-}L = \frac{1}{M} \sum_{i=1}^{m} \log p(w_i | w_{\backslash i}) \\
&\text{PPPL} = \exp(-\text{pseudo-}L)
\end{aligned}
\end{equation}

These metrics are based on the average token-level log probability, allowing for a fair evaluation across sentences of varying lengths in our experiments. For example, consider the comparison between \begin{CJK}{UTF8}{gbsn}他\end{CJK} (\textit{he}) and \begin{CJK}{UTF8}{gbsn}她自己\end{CJK} (\textit{herself}) in example~(\ref{minimalpair}). Although both correspond to one word in English, the former has fewer characters. Averaging sentence length helps mitigate the tendency of language models to favor shorter sentences \citep{song-etal-2022-sling}. Additionally, using perplexity as a common standard enables a more effective comparison of different language models' performance. We take the sentence in a sentence pair that has a lower perplexity as the models' preference. 

\subsection{Evaluation of closed-source LLMs}
For closed-source LLMs, i.e., \texttt{GPT-3.5-turbo} and \texttt{GPT-4o}, we use prompts to ask the model to select the more natural and acceptable sentence. The prompts can be found in Appendix~\ref{prompt}. 

\subsection{Human Evaluation}
To compare the performance of language models with humans, we recruited 24 native Mandarin speakers as volunteers to complete a cloze-filling task. To minimize bias toward any specific sentence structure, each participant annotated 5 sentences from each category, with sanity check sentences, totaling 70 sentences per person. Each sentence was annotated by three different participants, and the most frequently chosen response was adopted as the final annotation.

To assess the reliability of the annotations, we calculated Fleiss' Kappa for every group of three annotators. The average Fleiss' Kappa score reached 0.81\footnote{Most of the disagreement comes from the ambiguous setting where three native speakers might have different preferences.}, which indicates an ``almost perfect'' inter-annotator agreement \citep{Landis1977TheMO}.

\section{Result \& Discussion}
\subsection{Overall Result}

% \begin{figure*}[!th]
%     \centering
%     \includegraphics[width=\textwidth]{latex/mascfem.pdf}
%     \caption{Predicted Probabilities of Masculine and Feminine Pronouns in Different Experiments}
%     \label{fig:femmasc}
% \end{figure*}

% Unlike human evaluations, which achieve over 95\% accuracy in both synthetic and natural settings, language models show improved performance with natural data. Although natural language tends to include more distractors and longer sentences compared to synthetic data, models are better at capturing various constraints in natural contexts. This contrasts with the semantic parsing results noted by \cite{yang-schneider-2024-relative}. We hypothesize two possible reasons for this phenomenon: (1) natural data may better reflect the distribution of the training data, and (2) some sentences may appear in the training set.
% \section{Conclusion}
% Evaluating 21 language models across two data settings, our results reveal that none of the models consistently replicate human-like judgments. Notably, we observe that all language models rely more on their syntactic biases, even when modeling semantic constraints. The discrepancies in results between natural and synthetic data suggest that language models are better at modeling natural sequences but struggle to abstract the underlying complex linguistic constraints.

The results are summarized in Tables \ref{result_syn} and \ref{result_natural}, which present several noteworthy findings that we will discuss in detail in the following subsections. It is important to note that for the two closed-source LLMs, altering the order of the sentences within minimal pairs in the prompt significantly affected the results (see Appendix~\ref{gpt}). Therefore, we report the results with the sentences randomly shuffled in the minimal pairs. We also discuss the limitation of the prompt-based method in the Limitation section. Before diving into the detailed analysis, we would like to highlight a few key observations.

First, none of the models can match human performance in both settings. In the synthetic data setting,  \texttt{mengzi-bert-base} shows the best performance among all models, and in the natural setting, \texttt{glm-4-9b-chat} outperforms other models. Multilingual models perform worse than monolingual models.

Second, larger model sizes do not necessarily lead to better performance. In the synthetic data setting, \texttt{gpt2-distill} outperforms \texttt{gpt2-xlarge} with the same training data. Similarly, \texttt{chinese-pert-base} and \texttt{XLM-R-base} surpass their larger counterparts in both synthetic and natural settings. The two largest models \texttt{GPT-4o} and \texttt{GPT-3.5} show limited performance in this task as well.

As shown in Table~\ref{tab:average_performance}, the difficulty of various constraints is consistent across synthetic and natural data. However, all models perform better in the natural data setting, despite natural language often containing more distractors and longer sentences than synthetic data. This contrasts with the semantic parsing results noted by \citet{yang-schneider-2024-relative}.  We hypothesize two possible reasons for this phenomenon: (1) natural data may better reflect the distribution of the training data, suggesting that the models struggle to generalize the underlying abstract rules, and (2) the pretrained data is contaminated with our evaluation set. Most of our examples come from literature, making both hypotheses plausible for models trained on literary works or CommonCrawl. However, \texttt{bert-base-chinese}, \texttt{ernie-base}, and \texttt{mbert} are trained on data from Wikipedia and non-literary domains, indicating that the first hypothesis might be more likely\footnote{However, a recent study \citep{misra2024language} shows that language models can generalize rare phenomena from less rare ones. Validating this hypothesis requires rigorous experimental design, which we leave for future work.}. Additionally, we hypothesize that the second explanation applies to \texttt{GPT-4o}, given the significant difference in its performance between the synthetic and natural data.
\begin{table}
\small
    \centering
    \begin{tabular}{ccc}
    \toprule
    Binding & Syn Data & Natural Data\\
    \midrule
    Blocking &    41.3 & 65.5 \\
    Animacy & 87.5& 89.4 \\
    Verb$_{refl}$ & 40.2& 65.0 \\
   Verb$_{nonrefl}$ & 70.5& 74.8\\
   SO &  50.5& 69.6\\
   \bottomrule
    \end{tabular}
    \caption{Average accuracy of different binding phenomena across all evaluated models.}
    \label{tab:average_performance}
\end{table}

\begin{figure}
    \centering
    \includegraphics[width=\linewidth]{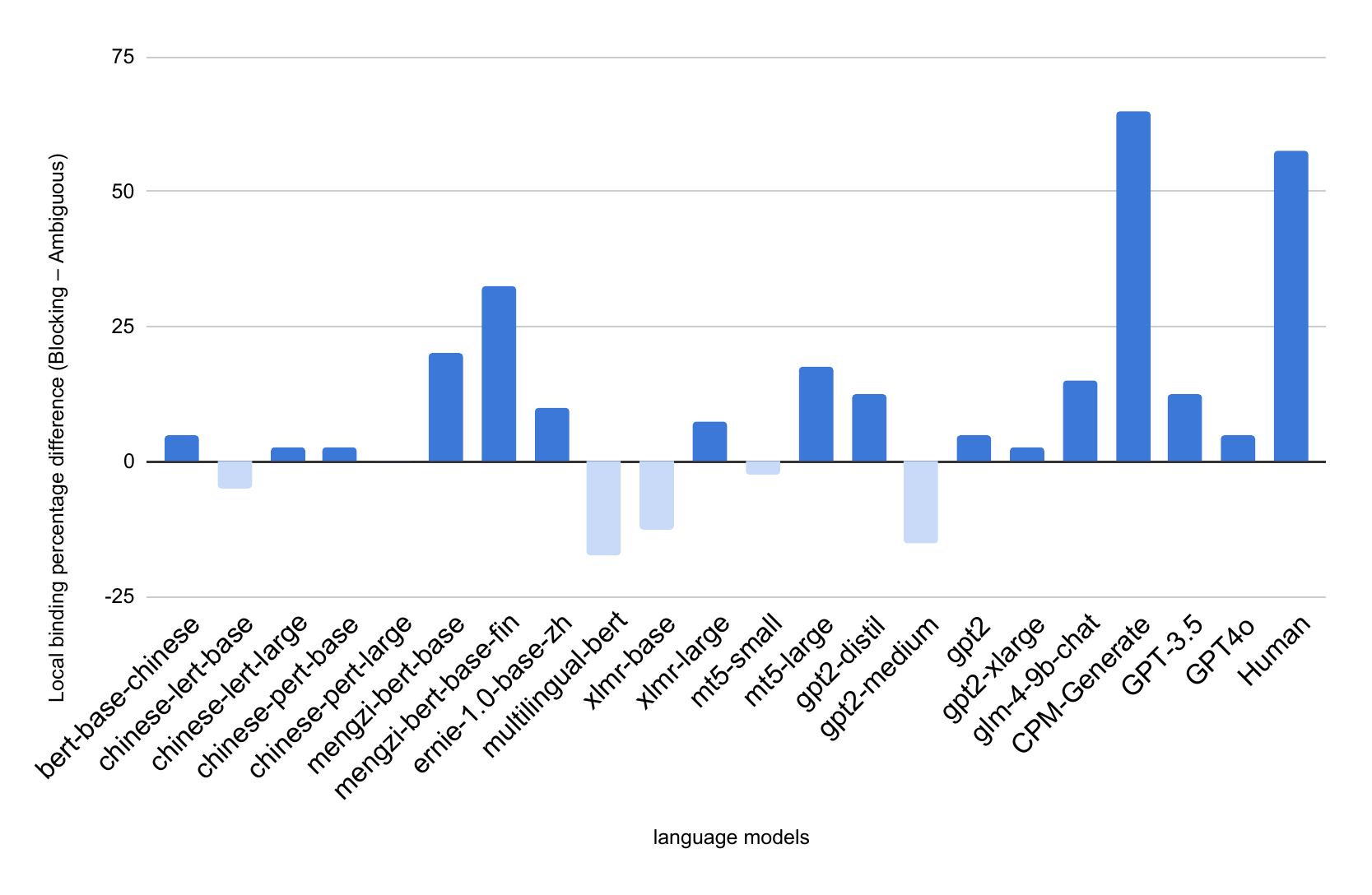}
    \caption{Local binding tendency caused by the blocking effect based on the baseline result.}
    \label{lb}
\end{figure}
\subsection{Language models show linear biases but not all language models prefer local binders}
 
% \begin{figure}
%     \centering
%     \includegraphics[width=1\linewidth]{local_binding.pdf}
%     \caption{Percentage of local binding across all models.}
%     \label{fig:lb}
% \end{figure}
Both \citet{song-etal-2022-sling} and \citet{xiang-etal-2021-climp} observe the models' vulnerability to linearly close distractors. Similar findings have been confirmed in other languages, such as English with \texttt{GPT-2} \citep{lee-schuster-2022-language} and Spanish with \texttt{mbert} \citep{de-dios-flores-etal-2023-dependency}. Our experiments show two key findings: (1) almost all languages have linear biases, yet (2) not all models show a bias toward local binders. In particular, most of the \textbf{encoder-only} models prefer \textbf{long-distance} binders while \textbf{decoder-only} models prefer \textbf{local binding} as shown in Table~\ref{result_syn}.

Regarding our first observation, we find that most models predict the blocking effect not because of the insertion of a first-person pronoun, but due to their linear preference. Since the examples in the Ambiguous Binding category are adapted from the Blocking Effect category by replacing first-person pronouns with third-person pronouns, this setup allows us to compare the results of these two groups and assess the influence of first-person pronouns on model behavior. Specifically, if language models have truly learned the underlying constraints of the blocking effect, they should assign higher probabilities to local binding readings when third-person pronouns are replaced with first-person pronouns. Consequently, we expect stronger local binding preferences in the blocking effect experiment than in the ambiguous binding experiment.

To quantify this, we define the local binding tendency as the difference between the number of local binding cases in the blocking effect and the number of local binding cases in the ambiguous binding category. Our findings reveal that most models -- except for \texttt{chinese-lert-base}, \texttt{xlmr-base}, \texttt{chinese-pert-base}, \texttt{gpt2-medium}, and \texttt{glm4} -- exhibit a slightly stronger tendency toward local binding in cases involving first-person pronouns, as illustrated in Figure~\ref{lb}. However, this tendency is generally weak across most models. The notable exception is \texttt{CPM-Generate}, which demonstrates a significant increase in its preference for local binding when a first-person pronoun is present, effectively mimicking human behavior in similar contexts. In conclusion, with the exception of \texttt{CPM-Generate}, all models appear to make correct predictions in the blocking effect setting primarily due to their linear bias, rather than an understanding of the constraints underlying the blocking effect pattern.

As for the second observation, we find that in the Blocking Effect and Verb Orientation (\textit{reflexive verbs}) settings, where \textit{ziji} should be bound to its local antecedent, most encoder-only models perform poorly. However, their performance improves in the Verb Orientation (\textit{non-reflexive verbs}) and Subject Orientation settings where long-distance binding is expected. In contrast, decoder-only models exhibit near-perfect performance in local-binding settings, such as the Blocking Effect, Animacy Effect, and Verb Orientation with Reflexive Verbs.

\subsection{Language Models Are More Sensitive to Semantics of Nouns than Verbs}
The animacy effect and two verb orientation experiments investigate whether language models possess the semantic knowledge required to resolve binding. As shown in the table, most models, except for \texttt{mt5-small}, perform well in the animacy setting, indicating they encode the knowledge that \textit{ziji} can only refer to an animate NP. This is particularly evident among the decoder-only models, which, despite exhibiting a strong bias toward local binding, can successfully switch to long-distance binding when the local binder is an inanimate noun, achieving nearly perfect accuracy.

\begin{figure}
\small
    \centering
    \includegraphics[width=\linewidth]{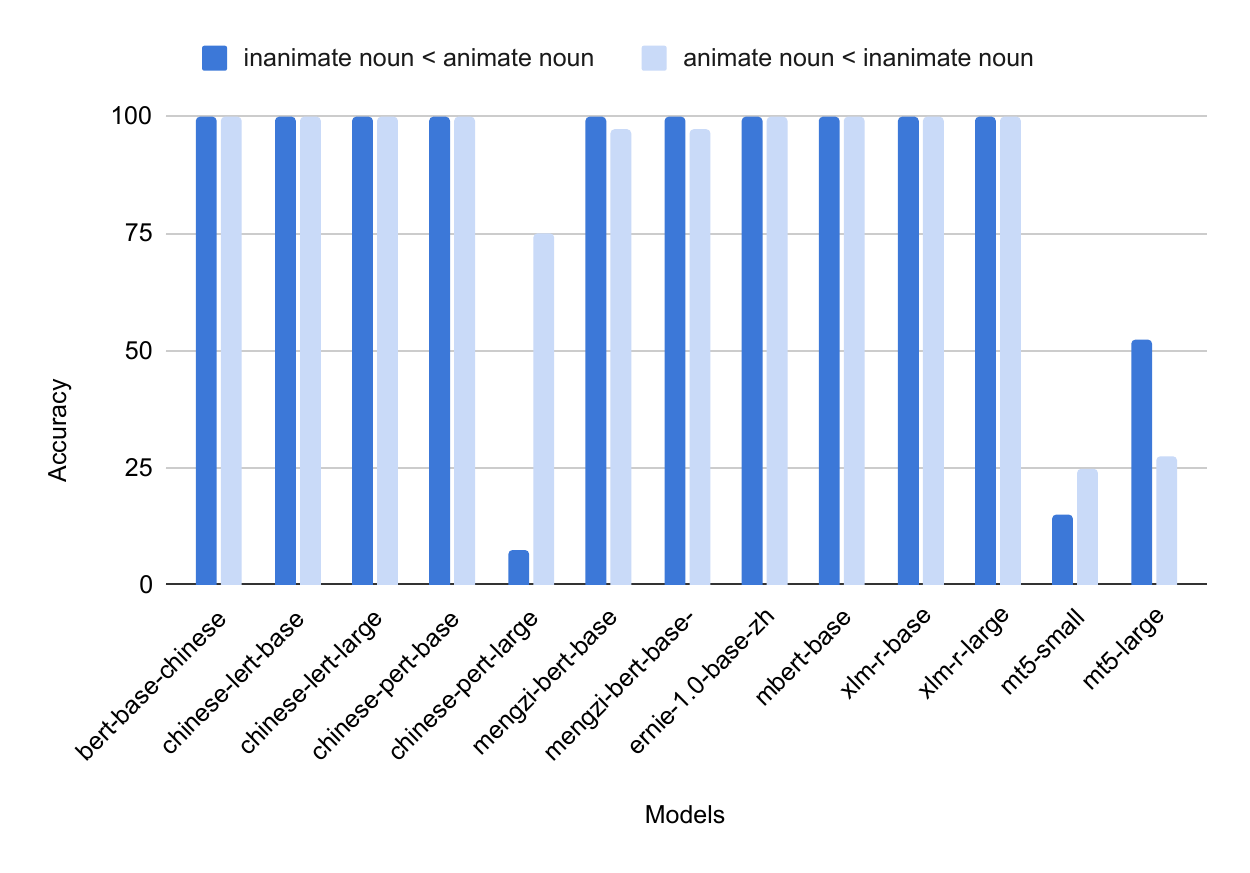}
    \caption{Accuracy of language models across two settings of the animacy effect: (1) matrix subject is animate and subordinate subject is inanimate (animate < inanimate), and (2) matrix subject is inanimate and subordinate subject is animate (inanimate < animate).}
    \label{fig:ani}
\end{figure}

This raises another question: is the success of the encoder-only models in the animacy setting due to their knowledge of animacy or their linear bias? To address this, we switch the order of the animate matrix subject and the inanimate subordinate subject, where local binding is the correct interpretation. As shown in Figure~\ref{fig:ani}, models that perform well in the typical animacy setting also excel in the switched experiment. This supports the first hypothesis: encoder-only models do learn animacy knowledge about \textit{ziji}. 

In contrast, none of the models perform equally well in the two Verb Orientation experiments which require different binding readings. We observe that models favoring local binding tend to perform poorly in non-reflexive verb scenarios, where the meanings of subordinate predicates necessitate lang-distance binding. Similarly, models preferring long-distance binding excel in non-reflexive verb settings but struggle with reflexive verbs. This pattern is particularly evident in synthetic data. 

Therefore, we argue that while most language models possess semantic knowledge of (animate) nouns, they struggle to understand the nuances of verb meanings. We assume that this difficulty may arise from the fact that the animacy of nouns is generally easier to distinguish than the reflexiveness of verbs. It is possible that there are more readily available distributional cues of animacy versus reflexiveness of verbs if we accept the assumption that all nouns can be either animate or inanimate while reflexive and non-reflexive verbs are less frequent than ambiguous verbs.

\section{Conclusion}
In this paper, we evaluated 21 language models across two data settings. Our results reveal that none of the models consistently replicate human-like judgments. We observe that all language models rely heavily on sequential biases, even when tasked with modeling syntactic and semantic cues. Furthermore, most models demonstrate a better understanding of the semantics of nouns compared to verbs. Several intriguing questions remain open. For instance, why do models find it easier to handle natural data, despite its longer sequences and more distractors, than synthetic data? Can language models generalize complex constraints based on more frequent and simpler linguistic phenomena? Why does SO not show a clear linear bias among LMs? We leave these questions for future research and welcome new insights into these areas.
\section*{Limitations}
We are aware that the prompt-based method for \texttt{GPT-3.5} and \texttt{GPT-4o} is \textbf{not directly comparable} to the perplexity-based approach, as results obtained using meta-linguistic prompts tend to perform worse than those derived from model representations \citep{hu-levy-2023-prompting}. As we mentioned in Appendix~\ref{gpt_perform}, both models are highly sensitive to the order of the sentence pairs in the prompt, with \texttt{GPT-3.5} showing a stronger bias toward Option-A. This observation remains consistent across different prompt designs. Therefore, the results we report may not fully reflect the language capability of these two LMs and our conclusion might not apply to them. We advise readers to interpret the results from these models with caution.

\section*{Ethical Considerations}
Our project had minimal computational costs since no additional model training was required. For human participants, informed consent was obtained prior to their participation in the questionnaire, and all collected data was anonymized and kept confidential to protect their privacy. Additionally, when creating and collecting sentences for the study, we ensured that the content was free from harmful or offensive material.

\section*{Acknowledgements}
We would like to express our gratitude to Amir Zeldes, Nathan Schneider, Tatsuya Aoyama, Wesley Scivetti, Yixiao Song, and all members of the NERT lab for their insightful suggestions and support. We are also deeply thankful to the volunteers who participated by completing the questionnaires; their contributions were essential to the success of this study. Finally, we extend our appreciation to the anonymous reviewers for their thoughtful and constructive feedback.
\bibliography{custom}
\newpage
\appendix
\section{In-Context Minimal Pair Templates for Different Binding Patterns}
\label{template}
    \begin{table}[!th]
\small
    \centering
    \begin{tabular}{ll}
    \toprule
    Constructions & In-context Minimal Pair Template \\
    \midrule 
    \parbox{1cm}{Blocking Effect} & \parbox{4cm}{\vspace{0.3cm}\textit{\textbf{Original:}\\}\begin{CJK}{UTF8}{gbsn}她\textsubscript{f}知道我\textsubscript{i}相信自己\textsubscript{i}。\end{CJK} \\ \textit{She\textsubscript{f} knows that I\textsubscript{i} trust myself\textsubscript{i}.}\\
    \textit{\textbf{Within Template}\\}\begin{CJK}{UTF8}{gbsn}如果她知道我相信自己， 那么我相信我自己/*她。\end{CJK} \\ \textit{If she\textsubscript{f} knows that I trust self, then I trust myself/*her.}
    \\}\\
    \midrule 
    
    Animacy Effect & \parbox{4cm}{\vspace{0.3cm}\textit{\textbf{Original:}\\}\begin{CJK}{UTF8}{gbsn}他\textsubscript{m}说这本书\textsubscript{t}改变了自己\textsubscript{m}。\end{CJK}  \\ 
    \textit{He\textsubscript{m} said the book changed him\textsubscript{m}.}\\
\textit{\textbf{Within Template}\\}\begin{CJK}{UTF8}{gbsn}如果他说这本书改变了自己，那么这本书改变了他/*它自己。\end{CJK}  \\
\textit{If he said that this book changed self, then this book changed him/*itself.}}\\
\midrule

    Subject Orientation & \parbox{4cm}{\vspace{0.3cm}\textit{\textbf{Original:}\\}\begin{CJK}{UTF8}{gbsn}他\textsubscript{m}给她\textsubscript{f}关于自己\textsubscript{f}的书。\end{CJK}  \\ 
    \textit{He\textsubscript{m} gave her his own book.}\\
\textit{\textbf{Within Template}\\}\begin{CJK}{UTF8}{gbsn}如果他给她关于自己的书，那么书是他/*她的。\end{CJK}  \\
\textit{If he gave her a book about self, then the book is about him/*her.}}\\
\midrule 
\parbox{1.5cm}{Verb Orientation}
    & \parbox{5cm}{\vspace{0.3cm}
    \textit{\textbf{Original:}}\\
    \begin{CJK}{UTF8}{gbsn}她\textsubscript{f}知道他\textsubscript{m}巴结自己\textsubscript{f/m}。\end{CJK} \\ \textit{She\textsubscript{f} knows that he\textsubscript{m} flatter her\textsubscript{f}/her\textsubscript{f}.}\\\textit{\textbf{Within Template:}\\}\begin{CJK}{UTF8}{gbsn}如果她知道他巴结自己，那么他巴结她/*他自己。\end{CJK}  \\
\textit{If she knows that he flattered self, he flattered her/*himself.}\\}\\
    \bottomrule
    \end{tabular}
    \caption{In-context minimal pair templates corresponding to various binding constructions.}
    \label{tab:template}
\end{table}
\section{Minimal Pair Examples for Natural Data}
\label{natural}
\pex Blocking Effect\\
\small
\a Original Sent:\\
\begin{CJK}{UTF8}{gbsn}她会第一个承认我真是有\textcolor{airforceblue}{自己}的一套习惯。
\end{CJK} \\
\begingl %% Start glosses
\gla ta\textsubscript{i} hui diyige chengren wo zhenshi you ziji de yitao xiguan //
\glb she would first admit I really have self DE one habit.//
\glft \textit{She would be the first to admit that I really have my own habit.}//
\endgl
\a Minimal pair sent I: \\
\begin{CJK}{UTF8}{gbsn}她会第一个承认我真是有\textcolor{airforceblue}{我自己}的一套习惯。
\end{CJK} \\
\begingl %% Start glosses
\gla ta\textsubscript{i} hui diyige chengren wo zhenshi you woziji de yitao xiguan //
\glb she would first admit I really have self DE one habit.//
\glft \textit{She would be the first to admit that I really have my own habit.}//
\endgl
\a Minimal pair sent II: \\
\begin{CJK}{UTF8}{gbsn}* 她会第一个承认我真是有\textcolor{airforceblue}{她自己}的一套习惯。
\end{CJK} \\
\begingl %% Start glosses
\gla ta\textsubscript{i} hui diyige chengren wo zhenshi you woziji de yitao xiguan //
\glb she would first admit I really have self DE one habit.//
\glft \textit{*She would be the first to admit that I really have her own habit.}//
\endgl
\xe
\pex Animacy Effect\\
\small
\a Original Sent:\\
\begin{CJK}{UTF8}{gbsn}...因为他还不懂得瘟疫在威胁着\textcolor{airforceblue}{自己}。
\end{CJK} \\
\begingl %% Start glosses
\gla ta\textsubscript{i} ... yinwei ta hai bu dongde wenyi zai weixie zhe ziji //
\glb ... because he yet NEG understand plague is threaten ASP   self.//
\glft \textit{... because he still doesn't understand that the plague is threatening him.}//
\endgl

\a Minimal pair sent I: \\
\begin{CJK}{UTF8}{gbsn}...因为他还不懂得瘟疫在威胁着\textcolor{airforceblue}{他自己}。
\end{CJK} \\
\begingl
\gla ... yinwei ta hai bu dongde wenyi zai weixie zhe ta-ziji //
\glb ... because he yet NEG understand plague is threaten ASP   himself.//
\glft \textit{... because he still doesn't understand that the plague is threatening him.}//
\endgl
\a Minimal pair sent II: \\
\begin{CJK}{UTF8}{gbsn}*...因为他还不懂得瘟疫在威胁着\textcolor{airforceblue}{它自己}。
\end{CJK} \\
\begingl
\gla ta\textsubscript{i} ... yinwei ta hai bu dongde wenyi zai weixie zhe ta-ziji //
\glb ... because he yet NEG understand plague is threaten ASP   itself.//
\glft \textit{*... because he still doesn't understand that the plague is threatening itself.}//
\endgl
\xe

\pex Verb Orientation Reflexive Verb\\
\small
\a Original Sent:\\
\begin{CJK}{UTF8}{gbsn}她会想，他在炫耀\textcolor{airforceblue}{自己}高人一等的教育。
\end{CJK} \\
\begingl %% Start glosses
\gla ta hui xiang, ta zai xuanyao ziji gaorenyideng de jiaoyu //
\glb she will think, he is boast self superior DE education //
\glft \textit{She would think, he is boasting about his superior education.}//
\endgl

\a Minimal pair sent I: \\
\begin{CJK}{UTF8}{gbsn}她会想，他在炫耀\textcolor{airforceblue}{他自己}高人一等的教育。
\end{CJK} \\
\begingl
\gla ta hui xiang, ta zai xuanyao ta-ziji gaorenyideng de jiaoyu //
\glb she will think, he is boast he-self superior DE education //
\glft \textit{She would think, he is boasting about his own superior education.}//
\endgl

\a Minimal pair sent II: \\
\begin{CJK}{UTF8}{gbsn}*她会想，他在炫耀\textcolor{airforceblue}{她自己}高人一等的教育。
\end{CJK} \\
\begingl
\gla ta hui xiang, ta zai xuanyao ta-ziji gaorenyideng de jiaoyu //
\glb she will think, he is boast she-self superior DE education //
\glft \textit{*She would think, he is boasting about her own superior education.}//
\endgl
\xe

\pex Verb Orientation Non-reflexive Verb\\
\small
\a Original Sent:\\
\begin{CJK}{UTF8}{gbsn}少女一下子注意到，少年正在目不转睛地望着\textcolor{airforceblue}{自己}。
\end{CJK} \\
\begingl %% Start glosses
\gla shaonv yixiazi zhuyidao, shaonian zhengzai mubuzhuanjing-de wang zhe ziji //
\glb girl suddenly notice, boy PROG fixedly-ADV gaze ASP self //
\glft \textit{The girl suddenly noticed that the boy was staring fixedly at her.}//
\endgl

\a Minimal pair sent I: \\
\begin{CJK}{UTF8}{gbsn}少女一下子注意到，少年正在目不转睛地望着\textcolor{airforceblue}{自己}。
\end{CJK} \\
\begingl
\gla shaonv yixiazi zhuyidao, shaonian zhengzai mubuzhuanjing-de wang zhe ziji //
\glb girl suddenly notice, boy PROG fixedly-ADV gaze ASP self //
\glft \textit{The girl suddenly noticed that the boy was staring fixedly at her.}//
\endgl

\a Minimal pair sent II: \\
\begin{CJK}{UTF8}{gbsn}*少女一下子注意到，少年正在目不转睛地望着\textcolor{airforceblue}{他自己}。
\end{CJK} \\
\begingl
\gla shaonv yixiazi zhuyidao, shaonian zhengzai mubuzhuanjing-de wang zhe ta-ziji //
\glb girl suddenly notice, boy PROG fixedly-ADV gaze ASP he-self //
\glft \textit{*The girl suddenly noticed that the boy was staring fixedly at himself.}//
\endgl
\xe

\pex Subject Orientation\\
\small
\a Original Sent:\\
\begin{CJK}{UTF8}{gbsn}王小姐带着马伯乐就到自己房里来。
\end{CJK} \\
\begingl %% Start glosses
\gla Wang xiaojie daizhe Ma Bole jiu dao ziji fang li lai //
\glb Miss Wang bring Ma Bole then arrive self room in come //
\glft \textit{Miss Wang brought Ma Bole and then went to her own room.}//
\endgl

\a Minimal pair sent I: \\
\begin{CJK}{UTF8}{gbsn}王小姐带着马伯乐就到她自己房里来。
\end{CJK} \\
\begingl
\gla Wang xiaojie daizhe Ma Bole jiu dao ta ziji fang li lai //
\glb Miss Wang bring Ma Bole then arrive she self room in come //
\glft \textit{Miss Wang brought Ma Bole and then went to her own room.}//
\endgl

\a Minimal pair sent II: \\
\begin{CJK}{UTF8}{gbsn}*王小姐带着马伯乐就到他自己房里来。
\end{CJK} \\
\begingl
\gla Wang xiaojie daizhe Ma Bole jiu dao ta ziji fang li lai //
\glb Miss Wang bring Ma Bole then arrive he self room in come //
\glft \textit{*Miss Wang brought Ma Bole and then went to his own room.}//
\endgl
\xe

\section{Prompts for GPT-3.5 and GPT-4o}
\label{prompt}
\textbf{Prompt A} \begin{CJK}{UTF8}{gbsn}下面两个句子哪个能自然，更容易接受？在这里，更自然指的是一个句子听起来符合母语者日常的语言使用习惯，读起来顺畅且易于理解。请只输出A或者B。A: 句子1 B: 句子2。
\end{CJK} \textit{Which of the following two sentences sounds more natural and is easier to accept? Here, ``more natural'' refers to a sentence that aligns with the everyday language use of native speakers, reads smoothly, and is easy to understand. Please output only ``A'' or ``B''. A: sentence 1 B: sentence 2.}
\\
\textbf{Prompt B} \begin{CJK}{UTF8}{gbsn}下面两个句子哪个能自然？请只输出A或者B, \textbf{然后给出解释}。A: 句子1 B: 句子2。
\end{CJK} \textit{Which of the following sentences sounds more natural? Please output only ``A'' or ``B'' \textbf{and then give me your explanations}. A: sentence 1 B: sentence 2.}
\\
\textbf{Prompt C} \begin{CJK}{UTF8}{gbsn}下面两个句子哪个能自然，更容易接受？在这里，更自然指的是一个句子听起来符合母语者日常的语言使用习惯，读起来顺畅且易于理解。请只输出A或者B, \textbf{然后给出解释}。A: 句子1 B: 句子2。
\end{CJK} \textit{Which of the following sentences sounds more natural and is easier to accept? Here, ``more natural'' refers to a sentence that aligns with the everyday language use of native speakers, reads smoothly, and is easy to understand. Please output only ``A'' or ``B'' \textbf{and then give me your explanations}. A: sentence 1 B: sentence 2.}
\\
\section{Performance of GPT-3.5 and GPT-4o with Varying Positions of the Correct Sentence in Prompts: Option A, Option B, or Mixed}
\label{gpt_perform}
This section presents the experimental results of \texttt{GPT-3.5} and \texttt{GPT-4o} tested by altering the order of sentence pairs, where the correct sentence is either always placed in Option A, always in Option B, or randomly shuffled between the two. Due to this bias, \texttt{GPT-3.5} does not perform well in the mixed setting either, because half the correct sentences in the sentence pairs are put in Option B.

As we can see, \texttt{GPT-3.5} shows a clear preference for Option A. When all correct sentences are placed in Option A in the prompt, \texttt{GPT-3.5} achieves perfect accuracy. However, when the correct sentences are all placed in Option B, its performance declines to the lowest accuracy.

Similarly, \texttt{GPT-4o} struggles to make consistent judgments when the order of the two sentences is switched, displaying a bias toward Option B instead.

The detailed results can be found in Table~\ref{table:gpt-perf} and Table~\ref{table:gpt-perf2}.

\begin{table*}[h!]
\label{gpt}
\centering
\begin{minipage}{0.5\textwidth}
    \centering
    \footnotesize
    \begin{tabular}{l|rrr|rrr}
    \toprule
    \textbf{} & \multicolumn{3}{c|}{\textbf{GPT-4o}} & \multicolumn{3}{c}{\textbf{GPT-3.5}} \\
    \toprule
    \textbf{Prompt} & \textbf{M} & \textbf{A} & \textbf{B} & \textbf{M} & \textbf{A} & \textbf{B} \\
    \toprule
    \textbf{Blocking} & 27.5 & 12.5 & 45.0 & 62.5 & 90.0 & 30.0 \\
    \textbf{Animacy} & 100.0 & 100.0 & 100.0 & 100.0 & 100.0 & 92.5 \\
    \textbf{Verb$_{refl}$} & 65.0 & 37.5 & 97.5 & 55.0 & 100.0 & 12.5 \\
    \textbf{Verb$_{nonrefl}$} & 100.0 & 97.5 & 100.0 & 57.5 & 100.0 & 20.0 \\
    \textbf{SO} & 50.0 & 30.0 & 77.5 & 50.0 & 100.0 & 0.0 \\
    \textbf{Average} & 68.5 & 55.5 & 84.0 & 65.0 & 98.0 & 31.0 \\
    \bottomrule
    \end{tabular}
    \caption{Performance of GPT-4o and GPT-3.5 across different order settings of the minimal pairs in the synthetic data setting. M: correct options have mixed orders; A: correction options are always option-A; B: correction options are always option-B}
    \label{table:gpt-perf}
\end{minipage}%
\hspace{0.5cm}
\begin{minipage}{0.5\textwidth}
    \centering
    \small
    \begin{tabular}{l|rrr|rrr}
    \toprule
    \textbf{} & \multicolumn{3}{c|}{\textbf{GPT-4o}} & \multicolumn{3}{c}{\textbf{GPT-3.5}} \\
    \toprule
    \textbf{Prompt} & \textbf{M} & \textbf{A} & \textbf{B} & \textbf{Mixed} & \textbf{A} & \textbf{B} \\
    \toprule
    \textbf{Blocking} & 100.0 & 97.5 & 95.0 & 57.5 & 90.0 & 25.0 \\
    \textbf{Animacy} & 97.5 & 100.0 & 95.0 & 80.0 & 97.5 & 72.5 \\
    \textbf{Verb$_{refl}$} & 100.0 & 100.0 & 100.0 & 60.0 & 97.5 & 22.5 \\
    \textbf{Verb$_{nonrefl}$} & 100.0 & 100.0 & 100.0 & 90.0 & 100.0 & 95.0 \\
    \textbf{SO} & 97.5 & 92.5 & 92.5 & 60.0 & 97.5 & 12.5 \\
    \textbf{Average} & 99.0 & 98.0 & 96.5 & 69.5 & 96.5 & 45.5 \\
    \bottomrule
    \end{tabular}
    \caption{Performance of GPT-4o and GPT-3.5 across different order settings of the minimal pairs in the natural data setting.}
    \label{table:gpt-perf2}
\end{minipage}
\end{table*}
\section{Training data distribution of evaluated language models}
The training data of the language models we test is listed in Table~\ref{model-training-data}.

\begin{table*}[th]
\centering
    \small
    \begin{tabular}{llll}
    \toprule
      Model   & Training Data Domain \\
    \midrule
      bert-base-chinese \citep{devlin-etal-2019-bert}          & Chinese Wikipedia   \\
      chinese-lert-base \citep{cui2022lert}     & Chinese Wikipedia, encyclopedia,
news, and question answering web  \\
      chinese-lert-large \citep{cui2022lert}    & Chinese Wikipedia, encyclopedia,
news, and question answering web  \\
      chinese-pert-base \citep{cui2022pert}    & Chinese Wikipedia, encyclopedia,
news, and question answering web  \\
      chinese-pert-large \citep{cui2022pert}    & Chinese Wikipedia, encyclopedia,
news, and question answering web  \\
      mengzi-bert-base \citep{zhang2021mengzi}     & Chinese Wikipedia, Chinese News, and
Common Crawl \\
      mengzi-bert-base-fin \citep{zhang2021mengzi}  & Chinese Wikipedia, Chinese News, and
Common Crawl, Finance data \\
      ernie-1.0-base-zh  \citep{sun2019ernie}   & Chinese Wikepedia, Baidu Baike,
Baidu news and Baidu Tieba\\
      mBERT \citep{devlin-etal-2019-bert}                &  Top 100 languages with the largest Wikipedias  \\
      XLM-R-base  \citep{Conneau2019UnsupervisedCR}          &  CommonCrawl \\
      XLM-R-large \citep{Conneau2019UnsupervisedCR}         &  CommonCrawl  \\
      \midrule
      \midrule
      mt5-small \citep{Xue2020mT5AM}            &  CommonCrawl\\  % Added a hypothetical value for illustration
      mt5-large  \citep{Xue2020mT5AM}            & CommonCrawl \\  % Added a hypothetical value for illustration
      \midrule
      \midrule
      GPT2 \citep{zhao2019uer}                & CLUECorpus-small (from Common Crawl) \citep{xu2020cluecorpus2020}  \\
      GPT2-medium \citep{zhao2019uer}          & CLUECorpus-small (from Common Crawl) \citep{xu2020cluecorpus2020}  \\
      GPT2-large  \citep{zhao2019uer}           & CLUECorpus-small (from Common Crawl) \citep{xu2020cluecorpus2020}  \\
      GPT2-xlarge \citep{zhao2023tencentpretrain}         & CLUECorpus-small (from Common Crawl) \citep{xu2020cluecorpus2020}  \\
      GLM-4-9b-chat \citep{glm2024chatglm}        & NA\\
      CPM-Generate \citep{zhang2021cpm}        & Encyclopedia, Webpage, Story, News, Dialog \\
      GPT-3.5 \citep{openai2023gpt35} & NA\\
      GPT-4o \citep{openai2023gpt4} & NA\\
         \bottomrule 
    \end{tabular}
    
    \caption{Models, training data and information source}
    \label{model-training-data}
 
\end{table*}
\end{document}